\documentclass[runningheads]{llncs}

\usepackage{eccv}

\usepackage{eccvabbrv}

\usepackage{graphicx}
\usepackage{booktabs}
\usepackage{colortbl}
\usepackage{multirow}
\usepackage{wrapfig}
\usepackage[accsupp]{axessibility}  

\usepackage{hyperref}

\usepackage{orcidlink}

\begin{document}

\title{COD: Learning Conditional Invariant Representation for Domain Adaptation Regression} 

\titlerunning{COD: Learning Conditional Invariant Representation for DAR}


\author{Hao-Ran Yang\inst{1} \and
Chuan-Xian Ren\inst{1}\thanks{Corresponding author.}\orcidlink{0000-0002-1861-3599} \and
You-Wei Luo\inst{1,2}\orcidlink{0000-0002-3027-6679} }

\authorrunning{H.~Yang et al.}

\institute{School of Mathematics, Sun Yat-Sen University, China \and
School of Mathematics, Jiaying University, China\\ 
\email{\{yanghr26@mail2, rchuanx@mail, luoyw28@mail2\}.sysu.edu.cn}}

\maketitle

\begin{abstract}
  Aiming to generalize the label knowledge from a source domain with continuous outputs to an unlabeled target domain, Domain Adaptation Regression (DAR) is developed for complex practical learning problems. However, due to the continuity problem in regression, existing conditional distribution alignment theory and methods with discrete prior, which are proven to be effective in classification settings, are no longer applicable. In this work, focusing on the feasibility problems in DAR, we establish the sufficiency theory for the regression model, which shows the generalization error can be sufficiently dominated by the cross-domain conditional discrepancy. Further, to characterize conditional discrepancy with continuous conditioning variable, a novel Conditional Operator Discrepancy (COD) is proposed, which admits the metric property on conditional distributions via the kernel embedding theory. Finally, to minimize the discrepancy, a COD-based conditional invariant representation learning model is proposed, and the reformulation is derived to show that reasonable modifications on moment statistics can further improve the discriminability of the adaptation model. Extensive experiments on standard DAR datasets verify the validity of theoretical results and the superiority over SOTA DAR methods.
\end{abstract}

\section{Introduction}
\label{sec:intro}

\begin{figure}[t]
  \centering
  \includegraphics[width=0.8\linewidth]{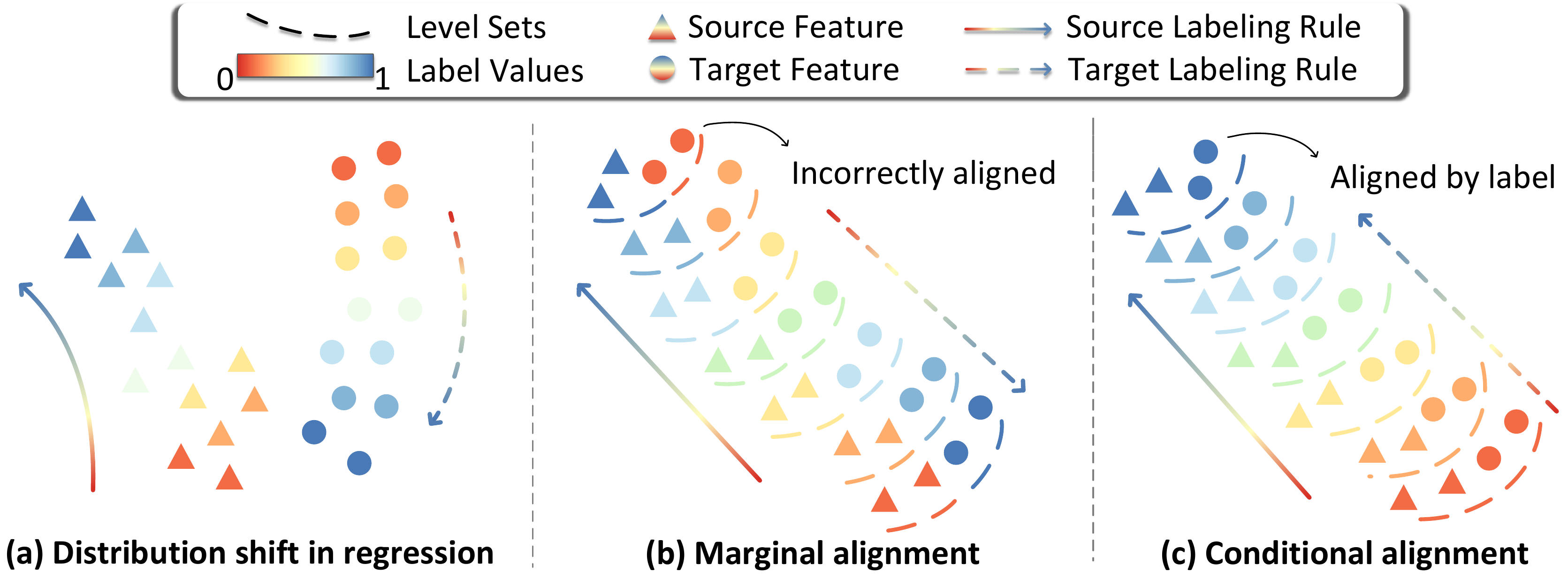}
  \caption{Illustration of the conditional shift in regression setting, where label value $y\in[0,1]$ and color gradients imply the continuity. \textbf{(a)} Before adaptation, distribution shift exists between source and target representations so the predictor trained on source domain cannot generalize to target domain. \textbf{(b)} After marginal distribution alignment, distributions of source and target domain are globally aligned. However, representations with different labels may be falsely aligned across domains, leading to significant inconsistency between cross-domain labeling rules. Thus, level sets provided by the source predictor are not suitable for target representations. \textbf{(c)} After conditional distribution alignment, label-wise matching is achieved, where the level sets of cross-domain labeling rules are consistent and the source predictor will be reliable. }
  \label{fig:sketch}
\end{figure}

Deep learning has shown a promising ability for fitting and achieved great success in various application tasks. However, large-scale data with sufficient annotations plays a vital role in training high-quality deep models while the label annotations for unlabeled data are usually expensive. Therefore, aiming to reduce training costs by transferring knowledge, Unsupervised Domain Adaptation (UDA)~\cite{ganin2016domain, long2015learning, gong2012geodesic, ben2010theory} has received increasing attention in recent years. Basically, UDA aims to transfer task-related information from a labeled source domain to a target domain without annotations, where the domain gap is characterized as the distribution shift across domains~\cite{quinonero2008dataset}. Following this line, considerable methods have been proposed to alleviate the UDA problem by shift correction and distribution alignment, such as adversarial training for cross-domain samples~\cite{ganin2016domain, tzeng2017adversarial} and explicitly distribution matching by statistical discrepancy minimization~\cite{long2015learning, sun2016return }. From a theoretical perspective, recent literature~\cite{zhao2019on, tachet2020domain} show that the marginal distribution $P_X$ with covariate shift assumption is insufficient to characterize the cross-domain generalization error, while the conditional distribution alignment indeed serves as a sufficient condition to ensure small error and successful transfer. Mathematically, when the conditional distributions $P_{X|y}$ for every $y$ are aligned, a well-trained source predictor can perform well on the target domain, which is described in~\cref{fig:sketch}. Inspired by theoretical advances, recent works~\cite{pei2018multi, saito2018maximum, kirchmeyer2021mapping} consider the statistical methods for conditional alignment, and show promising performance in mitigating negative transfer and improving recognition accuracy. However, most of the existing results are designed for classification settings with discrete prior on label variables, which induces the infeasibility for regression problems. Consequently, it restricts the application of UDA methods for wider real-world problems, \eg, human pose estimation~\cite{mao2022poseur}, product condition monitoring\cite{li2022novel}, depth estimation~\cite{peng2022rethinking}. 

Considering the dilemma of UDA, pioneer works have explored the Domain Adaptation Regression (DAR) framework for solving the distribution shift in continuous label scenarios, where theoretical results~\cite{mansour2009domain, cortes2011domain} and correction methods~\cite{geng2007automatic, teshima2020few, singh2020deep} are provided. To further relax the requirement of labeled training target data in pioneer DAR works, advanced works\cite{chen2021representation, nejjar2023dare} propose unsupervised DAR methods to learn transferable regression models without the help of labeled target data. Nonetheless, existing DAR theory and methods are directly facilitated by the covariate shift framework, where the marginal distribution alignment is considered. Therefore, it inevitably induces the negative transfer and misaligned local structure, which are shown in the results in classification settings~\cite{zhao2019on, tachet2020domain}. In fact, there is a significant obstacle for modeling conditional distribution alignment in regression scenarios, \ie, the conditional shift correction induces infinite matching problems for $P_{X|Y=y}$ on all points $y$. The continuous label variable with infinite possible outcomes makes it infeasible to directly apply the existing conditional alignment methods for DAR. Thus, it is vital to understand the continuity problem in regression, and develop the generalization error theory and statistical methodology for DAR scenarios.

In this work, we focus on the theory guarantee and conditional alignment methodology for successful DAR. Theoretically, we explore the generalization upper bound for continuous output scenarios and show the cross-domain joint error can be sufficiently upper-bounded by the conditional discrepancy. It ensures the sufficiency of conditional alignment still holds for regression scenarios and provides reliable guarantees for theory-driven modeling. Methodologically, to deal with the continuous problem in conditional alignment, we treat the conditional distribution as finite statistical moments in Hilbert space. By exploring the distribution embedding theory of Reproducing Kernel Hilbert Space (RKHS), a novel conditional metric called Conditional Operator Discrepancy (COD) is proposed, which can be interpreted as the conditional Wasserstein distance in RKHS. With the sufficiency theory and metric-based learning principle for regression, a COD-based conditional invariant representation learning method is proposed, which aims to correct the conditional shift on the pushforward measures in latent representation space. By reformulating the empirical COD estimation, a modified principle is developed to further improve the discriminability during knowledge transfer. Generally, our contributions can be summarized as follows:
\begin{itemize}
    \item The sufficiency theory for the conditional shift framework is established for DAR settings, which shows the cross-domain conditional discrepancy on continuous label variable serves as the major term in generalization error.
  \item A novel COD metric for characterizing conditional distribution discrepancy with continuous and infinite conditioning variable is explored, providing a feasible solution for conditional discrepancy optimization in DAR. 
  \item A COD-based representation learning model with reformulation on moment statistics is proposed, which admits discriminability on task-related knowledge and shows superior performance in extensive experiments.
\end{itemize}

\section{Related Work}

\noindent
\textbf{Domain Adaptation Classification.}
Based on the distribution shift assumption, most of the UDA methods can be roughly categorized into three types: methods that align marginal distribution, conditional distribution and joint distribution. Most UDA methods consider only the marginal distribution alignment. Discrepancy-based methods attempt to directly minimize the distribution discrepancy and different types of discrepancy measurement greatly enrich this type of methodology.
For example, these methods adopt moment matching~\cite{long2015learning, sun2016return } or extension under manifold framework~\cite{gong2012geodesic, luo2020unsupervised}. In particular, Optimal transport (OT) and Wasserstein distance are introduced to UDA as a sound measurement for domain shift, based on which a large number of methods have been proposed~\cite{courty2014domain, courty2017joint, damodaran2018deepjdot, shen2018wasserstein, lee2019sliced}. Zhang~\etal~\cite{zhang2019optimal} further develop both theory and methodology to RKHS and Luo \etal ~\cite{luo2021conditional} extends Kernel Bures metric to conditional distribution alignment. Adversarial-based methods~\cite{ganin2016domain, hoffman2018cycada, hu2022learning} learn domain-invariant features via two-player games. Specifically, they alternatively optimize the feature generator and the domain discriminator to achieve domain confusion. In order to obtain better performance, advanced methods have explored how to utilize the label information to achieve conditional distribution alignment~\cite{pei2018multi, saito2018maximum, Zhao2020Conditional, tang2020discriminative, wang2020classes, chen2022reusing, ren2022buresnet, luo2023mot} or joint distribution alignment~\cite{long2017deep, long2018conditional }. However, UDA methods for shift correction are proposed for discrete and finite conditioning variables, which are inapplicable for regression.

\noindent
\textbf{Domain Adaptation Regression.}
DA for regression has received relatively little attention compared with classification. 
Mansour \etal~\cite{mansour2009domain} and Cortes \etal~\cite{cortes2011domain} introduce theoretical analysis for DA in regression scenarios. Zhao \etal~\cite{zhao2022fundamental} discuss the tradeoffs between accuracy and invariance in learning invariant representations in both classification and regression settings. 
A range of methods in the shallow regime are proposed to tackle the DAR problem. These methods can be roughly categorized into two types: 
importance weighting~\cite{geng2007automatic, de2021adversarial} and learning domain-invariant representation~\cite{ pan2010domain, courty2017joint, nikzad2020domain}. Besides, some attempts have been made in deep regime to tackle regression problem~\cite{teshima2020few, singh2020deep, li2021learning, wei2022adaptive}. Unfortunately, most of them belong to semi-supervised DA, which means they need to use a small number of target domain labels. Recently, deep UDA methods for regression that get rid of target label are proposed~\cite{chen2021representation, nejjar2023dare, wu2022distribution, dhaini2023unsupervised}. 
RSD~\cite{chen2021representation} states that deep neural networks are less robust to feature scaling in regression problems and aligning the deep features directly will impede task performance for altering feature scale. Thus, a strategy of aligning orthogonal bases of both domains is adopted in RSD. 
Based on RSD, DARE-GRAM~\cite{nejjar2023dare} aligns the inverse Gram Matrix of deep features of both domains, and ensures promising improvements in accuracy and stability. Meanwhile, several DAR methods for specific tasks have been explored~\cite{jiang2021regressive, wang2022contrastive, bao2022generalizing, goswami2022deep, li2022domain, li2022novel}. However, the theory and method for conditional shift correction are still unexplored for DAR scenarios, which hinders the in-depth reasoning and development for practical modeling.

\section{COD: Theory and Methodology}

\subsection{Preliminary}
\label{sec:preliminary}

\textbf{Setup}.
In this paper, we discuss UDA method dealing with regression problems. Let X and Y be covariate and label variable taken values on $\mathcal{X}$ and $\mathcal{Y}$. $\mathbb{P}$ denotes the set of distributions and subscripts represent the corresponding random variable while the superscript represents the domain, \eg $P_{XY}^s \in \mathbb{P}_{\mathcal{XY}}$ implies the joint distribution on source domain. 
In UDA settings, we are provided with labeled source samples $(x^s, y^s) \sim P_{XY}^s$ and unlabeled target samples $(x^t, y^t) \sim P_{XY}^t$, which means the target labels are not available during training. Samples of source and target domain follow different distributions,~\ie $P_{XY}^s \neq P_{XY}^t$. Denote $g: X\to Z$ as the representation extractor, $h: Z \to Y$ as the predictor, $\varepsilon_s$ and $\varepsilon_t$ as the source and target prediction error, respectively. UDA methods aim to generalize the learner $h{\circ}g$ across domains and minimize target risk $\varepsilon_t$ by mitigating the distribution shift problem. In contrast to the discrete label space in classification settings, unsupervised DAR discusses UDA problems with continuous labels, which means the number of classes is uncountable, leading to great difficulty in aligning conditional distribution $P_{X|Y}$. \textit{The proofs are provided in supplementary materials.}

\noindent
\textbf{OT in RKHS.}
Let $(\mathcal{X},\mathcal{B})$ be a measure space with Borel $\sigma-$field $\mathcal{B}$. 
Denote $(\mathcal{H}_{\mathcal{X}},k_{\mathcal{X}})$ as the RKHS of $\mathcal{X}$, 
which is generated by the positive definite kernel $k_{\mathcal{X}}$. 
Elements from $\mathcal{X}$ are mapped to $\mathcal{H}_{\mathcal{X}}$ via feature map $\phi(x)=k_{\mathcal{X}}(x,\cdot)$, which is assumed to satisfy the reproducing properties $\left<\phi(x),\phi(x')\right>_{\mathcal{H}_{\mathcal{X}}}=k_{\mathcal{X}}(x,x')$ and $\left<\phi(x),f\right>_{\mathcal{H}_{\mathcal{X}}}=f(x),\forall f\in\mathcal{H}_{\mathcal{X}}$.
The mean element $\mu_X$ in $\mathcal{H}_{\mathcal{X}}$ with law $P_X$ is given by $\mu_X=\mathbb{E}_X\left[\phi(X)\right]$, and the covariance operator $\mathcal{C} _{XX}$ is defined as $\mathcal{C} _{XX} = E_{X}(\phi(X)-\mu_X)\otimes(\phi(X)-\mu_X)$.

For any two distributions $P_X^s,P_X^t\in \mathbb{P}_{\mathcal{X}}$, the Kantorovitch optimal transport in RKHS $\mathcal{H}_{\mathcal{K}}$ is considered on pushforward measures $\phi_{\#} P_X^s$ and $\phi_{\#} P_X^t$:
\begin{align}
  \label{fuc:kernel_OT}
  d^2_\mathrm{KW}(X^s, X^t) = \inf_{\pi_{\mathcal{K}}\in\Pi(\phi_{\#}P_X^s,\phi_{\#}P_X^t)}\int_{\mathcal{H}_{\mathcal{K}}\times\mathcal{H}_{\mathcal{K}}}\left\|X^s-X^t\right\|_{\mathcal{H}_{\mathcal{K}}}^2d\pi_{\mathcal{K}}(X^s,X^t)
\end{align}
Zhang \etal~\cite{zhang2019optimal} show that if the pushforward measures are Gaussian, the Kernel Gaussian Wasserstein (KGW) distance between them can be written as
\begin{align}
  \label{fuc:KGW}
  d_\mathrm{KGW}^2 (X^s, X^t) & =  \| u^s _{X} - u^t _{X} \|^2_{\mathcal{H} _\mathcal{K}} + tr(\mathcal{C}^{ss} _{XX} + \mathcal{C}^{tt} _{XX} - 2\mathcal{C}^{st} _{XX}) 
\end{align}
where 
$\mathcal{C}^{st} _{XX}  = \sqrt{\sqrt{\mathcal{C}^{ss} _{XX}}\mathcal{C}^{tt} _{XX}\sqrt{\mathcal{C}^{ss} _{XX}}}$
and $\mathcal{C}^{ss} _{XX}$, $\mathcal{C}^{tt} _{XX}$ are the covariance operators of $P_X^s$ and $P_X^t$ on $\mathcal{H}_{\mathcal{K}}$, respectively. 
\begin{remark}
  Note that KGW consists of two terms, \ie, the Maximum Mean Discrepancy (MMD) on first-order statistics and kernel Bures metric on second-order statistics. It provides a basic innovation for exploring conditional discrepancy metric by considering the conditional statistics in RKHS, \ie, first-order statistics in literature~\cite{song2009hilbert} and second-order statistics in literature~\cite{fukumizu2009kernel}.
\end{remark}

\subsection{Generalization Error Analysis for DAR}
Most of DAR methods focus on aligning the marginal distribution $P_X$. However, Zhao~\etal~\cite{zhao2019on} have shown that in the presence of label shift, only aligning the marginal distribution does not guarantee satisfying generalization performance on target domain. In fact, \cite{zhao2019on, tachet2020domain} have proven the following theorem~\eqref{them:lower_bound} for classification problems:
\begin{theorem}[\cite{zhao2019on}]
  \label{them:lower_bound}
  For representation variable $Z = g(X)$, suppose inequality  $d_{JS}(P_Y^s,P_Y^t) \geq d_{JS}(P_Z^s,P_Z^t)$ holds, then for predictor $h: Z \to Y$, we have
  \begin{align}
    \varepsilon_s(h{\circ}g){+}\varepsilon_t(h{\circ}g) {\geq} \frac12 \left(\sqrt{d_{JS}(P_{Y}^s,P_{Y}^t)} - \sqrt{d_{JS}(P_{Z}^s,P_{Z}^t)} \right)^{2}, 
  \end{align}
  where $d_{JS}$ denotes the Jensen-Shannon divergence.
\end{theorem}

Assuming the label distributions differ between source and target domain, \ie, $d_{JS}(P_Y^s,P_Y^t) > 0$, the lower bound shows that good alignment of representation distributions leads to bad joint error. In other words, when label shift exists, which is rather common in realistic scenarios, only aligning the marginal distribution is not enough for low target error.

To deal with the insufficiency problem shown in the lower bound analysis, it is intuitive to introduce label information into generalization analysis, \eg, the balanced error rate for classification scenarios~\cite{tachet2020domain}. Analogously, for the continuous regression scenarios, we first define the continuous balanced error rate for quantifying maximum conditional risk in DAR.
\begin{definition}
  \label{def:BER_con}
  The continuous balanced error rate of predictor $h$ on distribution $P$ is defined as:
  \begin{align}
    \mathrm{cBER}_\mathrm{P}(h,Y):=\sup_{(x,y) \in \mathcal{X} \oplus \mathcal{Y} }P(h(x) \neq Y\mid Y=y).
  \end{align}
\end{definition}
Intuitively, the continuous balanced error rate can be regarded as the maximum potential risk for predicting sample $x$ with label $y$. Based on $\mathrm{cBER}$ and conditional shift correction with representation transformation $Z=g(X)$, a sufficient condition for DAR can be induced as follows.
\begin{theorem}
  \label{them:sufficient_con}
  For representation variable $Z^s = g(X^s)$ and $Z^t = g(X^t)$, if $g(\cdot)$ satisfies conditional invariant property, \ie, 
  \begin{equation}
        \begin{aligned}
    P_{Z|Y = y}^s = P_{Z|Y = y}^t, ~~ \forall y \in \mathcal{Y},
  \end{aligned}
  \end{equation}

  then for any predictor $h: Z \to Y$, we have 
  \begin{align}
    \varepsilon_{s}(h)+\varepsilon_{t}(h)\leq2\mathrm{cBER}_\mathrm{P^{s}}(h, Y).
  \end{align}
\end{theorem}

Therefore, the key for DA, no matter for classification or regression problems,  is to achieve conditional distribution alignment. In classification settings, conditional alignment can be achieved by conducting class-wise marginal distribution alignment with the help of target pseudo label. However, in regression scenarios, the label variable is continuous, which means previous methods correcting conditional shift in classification settings are no longer applicable. Obviously, it is impossible to align the $P_{X|y}$ for every single $y$ when $y$ is a continuous variable and the possible number of y is infinite. Thus, it is important and meaningful to explore a new framework that is capable of conditional distribution alignment for continuous label variable.

\subsection{COD metric for DAR}

\noindent
\textbf{Theory for COD}.
Due to the continuity of label variable, it is impractical to correct conditional shift by aligning marginal distribution of $P_{X|y}$ for every single $y$. For this inevitable obstacle, an intuitive idea is can we align all $P_{X|y}$ as a whole? \ie, $P_{X|Y}$, rather than align $P_{X|y}$ one by one? It is more than attractive that if we can characterize $P_{X|Y}$ with a well-defined discrepancy, then all $P_{X|y}$ can be aligned by aligning the whole $P_{X|Y}$.  

In kernel embedding theory, a distribution can be characterized by a point in RKHS. Then $P_{X|Y}$ can be seen as a point set in RKHS in which every element corresponds to a distribution $P_{X|y}$ for a fixed $y$. 
From another perspective, we can consider $P_{X|Y}$ as the distribution in RKHS and measure the cross-domain conditional discrepancy by defining the COD distance on their statistical moments in RKHS. Following the definitions of conditional statistics in pioneer works, \ie, the conditional mean operator as $\mathcal{U}_{X|Y}:=\mathcal{C}_{XY}\mathcal{C}_{YY}^{-1}$~\cite{song2009hilbert} and conditional covariance operator as $\mathcal{C}_{XX|Y}=\mathcal{C}_{XX}-\mathcal{C}_{XY}\mathcal{C}_{YY}^{-1}\mathcal{C}_{YX}$~\cite{fukumizu2009kernel}, we present the definition of COD as follows. 
\begin{definition}[COD] 
Given conditional distributions $P_{X|Y}^s$ and $P_{X|Y}^t$, denote $\mathcal{C}^{ss} _{XX|Y}$ and $\mathcal{C}^{tt}_{XX|Y}$ as their corresponding covariance operators. The COD is defined as 
  \begin{equation}
    \begin{aligned}
       d_\mathrm{COD}^2 ({P} _{X|Y}^s, {P} _{X|Y}^t) = & \| \mathcal{U}^s _{X|Y} - \mathcal{U}^t _{X|Y} \|^2_{\mathcal{H} _\mathcal{K}} 
       + \mathrm{tr}(\mathcal{C}^{ss} _{XX|Y} + \mathcal{C}^{tt} _{XX|Y} - 2\mathcal{C}^{st} _{XX|Y}),
    \end{aligned}
  \end{equation}
  where $\mathcal{C}^{st} _{XX|Y} = \sqrt{\sqrt{\mathcal{C}^{ss} _{XX|Y}}\mathcal{C}^{tt} _{XX|Y}\sqrt{\mathcal{C}^{ss} _{XX|Y}}}$.
\end{definition}

\begin{figure}[tb]
  \centering
    \includegraphics[width=0.95\linewidth]{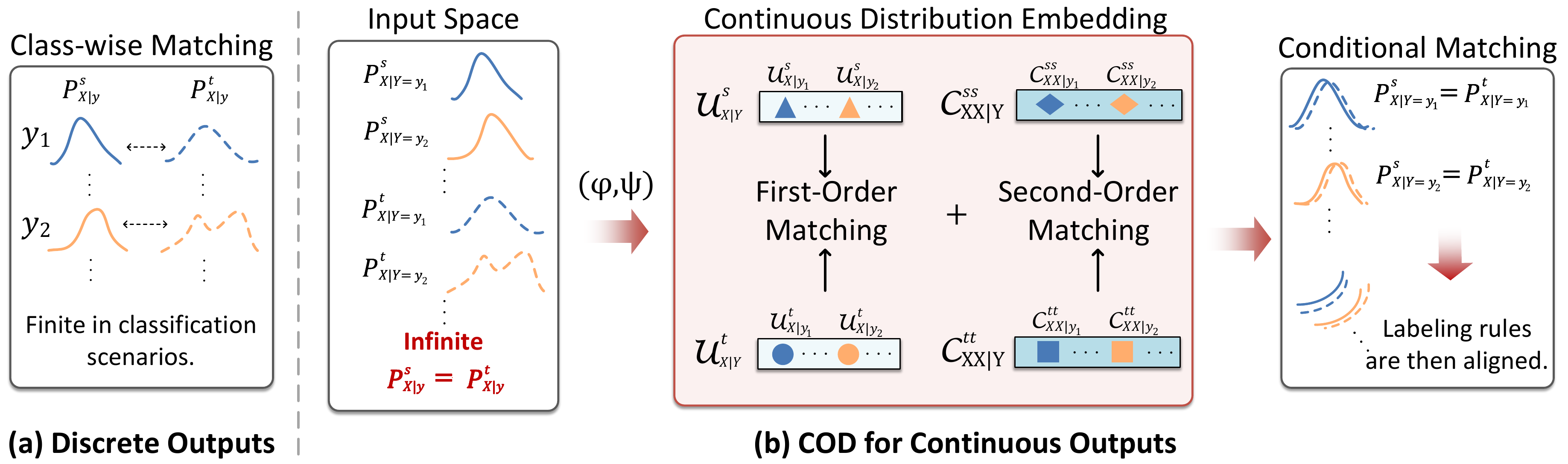}
  \caption{Illustration of COD metric. \textbf{(a)} Classification DA methods usually correct conditional shift by class-wise computations on discrete clusters $P_{X|y}$, which is infeasible for regression due to the infinite slices $P_{X|y}$. \textbf{(b)} In COD, the continuous conditional distributions are embedded into the RKHS and characterized by finite statistical moments in kernel spaces, which do not rely on specific conditions $y$. Under the guarantees of distribution embedding property, the conditional alignment is equivalent to the matching on first-order and second-order statistics, \ie, mean embedding operator $\mathcal{U}_{X|Y}$ and covariance operator $\mathcal{C}_{XX|Y}$. When COD is minimized, conditional alignment is achieved, and the level sets of cross-domain labeling rules are then aligned. 
  }
  \label{fig:COD}
\end{figure}

Theoretically, it can be proved that COD indeed admits the metric property on conditional distributions, \ie, two conditional distributions are the same if and only if the COD metric between them is zero. An intuitive illustration of COD is shown in~\cref{fig:COD}. Formally, COD admits two key properties.
\begin{proposition}
  (a) COD defines a metric between Gaussian measures.\\
  (b) Let $(\mathcal{X},\mathcal{B}_{\mathcal{X}})$ be the locally compact and Hausdorff measurable space and k be $c_0$-universal kernel. 
  Assuming that $(\phi(X),\psi(Y))$ is a Gaussian random variable in $\mathcal{H}_{\mathcal{X}}\oplus\mathcal{H}_{\mathcal{Y}}$. 
  For any $P_{X|Y}^{s},P_{X|Y}^{t}\in\mathbb{P}_{\mathcal{X}|\mathcal{Y}}$, we have   
  \begin{equation}
    \begin{aligned}
    d_\mathrm{COD}^2(P_{X|Y}^{s},P_{X|Y}^{t}) = 0 \implies P_{X|Y}^{s} = P_{X|Y}^{t}.
  \end{aligned}
  \end{equation}

\end{proposition}

\noindent
\textbf{Empirical Estimation of COD}.
Based on the above properties, we can perform conditional distribution alignment by minimizing the COD metric between source and target domain. To do this, we shall derive the empirical estimation of COD, which can be obtained with the help of the kernel trick.

Let ${(\mathbf{x}_i^s, \mathbf{y}_i^s)}_{i=1}^{n_s}$ and ${(\mathbf{x}_i^t, \mathbf{y}_i^t)}_{i=1}^{n_t}$ be two set of samples drawn i.i.d. from source and target domain. For simplicity, $n_s$ and $n_t$ are both set to $n$. In kernel method, samples are mapped to RKHS $\mathcal{H}_{\mathcal{X}} \oplus \mathcal{H}_{\mathcal{Y}}$ by the implicit feature map $(\phi, \psi)$. Denote the feature map matrices by $(\mathbf{\Phi}^s, \mathbf{\Psi}^s)$ and $(\mathbf{\Phi}^t, \mathbf{\Psi}^t)$, the explicit kernel matrices $\mathbf{K}^{ss}_{X} = \mathbf{\Phi}^s{\mathbf{\Phi}^s}^T $, $\mathbf{K}^{ss}_{Y} = \mathbf{\Psi}^s{\mathbf{\Psi}^s}^T$ are computed as $(\mathbf{K}^{ss}_{X})_{ij} = k_{\mathcal{X}}(\mathbf{x}_i^s,\mathbf{x}_j^s)$, $(\mathbf{K}^{ss}_{Y})_{ij} = k_{\mathcal{Y}}(\mathbf{y}_i^s,\mathbf{y}_j^s)$, respectively. And so as $\mathbf{K}^{tt}_{X}$, $\mathbf{K}^{st}_{X}$ and $\mathbf{K}^{tt}_{Y}$. Let $\mathbf{I}_n$ be the n-dimensional identity matrix, $\mathbf{1}_n$ be the n-dimensional vector with all elements equal to 1. The $n$-dimensional centering matrix can be defined as $\mathbf{H}_n=\mathbf{I}_n-\frac1n \mathbf{1}_n \mathbf{1}_n^T$. Then the centralized kernel matrix is defined as $\mathbf{G} = \mathbf{H}_n \mathbf{K} \mathbf{H}_n$.

With above notations, the covariance operator $\mathcal{C}_{XY}$ can be empirically  estimated as $\hat{\mathcal{C}}_{XY}=\frac{1}{n}(\mathbf{\Phi}-\hat{\mu}_{X} \mathbf{1}^{T})(\mathbf{\Psi}-\hat{\mu}_{Y} \mathbf{1}^{T})^{T}=\frac{1}{n}\mathbf{\Phi} \mathbf{H}\mathbf{\Psi}^{T}$, then the conditional moment statistics can be estimated as
\begin{equation}
    \begin{aligned}
      \hat{\mathcal{U}} _{X|Y} = (\frac{1}{n}\mathbf{\Phi} \mathbf{H}\mathbf{\Psi}^{T})(\frac{1}{n}\mathbf{\Psi} \mathbf{H}\mathbf{\Psi}^{T})^{-1}, ~~~
       \hat{\mathcal{C}}_{XX|Y}=\frac{1}{n}{\mathbf{\Phi}}\mathbf{H}_n{\mathbf{A}}\left({\mathbf{\Phi}}\mathbf{H}_n{\mathbf{A}}\right)^T ,
    \end{aligned}
\end{equation}
where $\mathbf{A}$ satisfies that $\mathbf{B}={\mathbf{A}}{{\mathbf{A}}}^T$ and $B \triangleq \mathbf{I}_{n}-\frac{1}{n\varepsilon}\left[{\mathbf{G}}_{Y}-{\mathbf{G}}_{Y}\left({\mathbf{G}}_{Y}+\varepsilon n{\mathbf{I}}_{n}\right)^{-1}{\mathbf{G}}_{Y}\right]$. Based on these estimations, the closed-form empirical estimation of the COD metric can be derived as follows.
\begin{proposition}
  The empirical Conditional Operator Discrepancy distance is 
  \begin{equation}
    \label{fuc:emp_COD}
    \begin{aligned}
      \widehat{d}_\mathrm{COD}^2 =&\ \mathrm{tr}(\mathbf{K}^{ss}_{Y}(\mathbf{K}^{ss}_{Y})^{-1}\mathbf{K}^{ss}_{X}(\mathbf{K}^{ss}_{Y})^{-1})  + \mathrm{tr}(\mathbf{K}^{tt}_{Y}(\mathbf{K}^{tt}_{Y})^{-1}\mathbf{K}^{tt}_{X}(\mathbf{K}^{tt}_{Y})^{-1}) \\  
      &-2\mathrm{tr}(\mathbf{K}^{ts}_{Y}(\mathbf{K}^{ss}_{Y})^{-1}\mathbf{K}_{X^sX^t}(\mathbf{K}^{tt}_{Y})^{-1}) \\ 
      & + \varepsilon\mathrm{tr}\left[{\mathbf{G}}_X^s\left(\varepsilon n{\mathbf{I}}_n+{\mathbf{G}}_Y^s\right)^{-1}\right]+\varepsilon\mathrm{tr}\left[\mathbf{G}_X^t\left(\varepsilon n{\mathbf{I}}_n+\mathbf{G}_Y^t\right)^{-1}\right] \\ 
      & - \frac2{n}\left\|\left(\mathbf{H}_n{\mathbf{A}}^t\right)^T{\mathbf{K}}^{ts}_{X}\left(\mathbf{H}_n{\mathbf{A}}^s\right)\right\|_*, 
    \end{aligned}
  \end{equation}
  where $\left\| \cdot \right\|_*$ is the nuclear norm. Empirically, $\mathbf{K}_{Y}^{-1}$ is approximated by the regularized formulation $(\mathbf{K}_{Y}+\lambda \mathbf{I})^{-1}$. 
\end{proposition}

\noindent
\textbf{Discriminability of empirical COD}.
As discussed above, the first term of COD measures the distance of the first-order statistic between conditional distributions in RKHS. However, recent study~\cite{chen2019graph, wang2021rethinking} shows that the mean matching may reduce the discriminability when bringing the two distributions close, which undoubtedly degrades the performance of the algorithm. Therefore, it is worth making an in-depth analysis of the discriminability of COD as well.   

Without loss of generality, we first consider Kronechker Delta kernel on conditioning variable $Y$, \ie, $k(\mathbf{y}_1,\mathbf{y}_2) = \delta(\mathbf{y}_1,\mathbf{y}_2)$.
More explicitly, $k(\mathbf{y}_1,\mathbf{y}_2)$ equals to $1$ if $\mathbf{y}_1=\mathbf{y}_2$ or $0$ otherwise. For empirical scenarios with finite samples, the numbers of samples and observed labels are finite, which implies the number of distinct label $c$ will be obviously no more than sample size $n$. For convenience, we call two samples are in the same class when their label values are the same. Consequently, the first term in COD can be reformulated as follows.
\begin{proposition}
  Under the conditions mentioned above, the empirical estimation of $d_{CMMD}$ can be written as 
  \begin{equation}
    \label{fuc:emp_CMMD}
    \begin{aligned}
      \widehat{d}_\mathrm{CMMD}^2 = 
      & \sum_{p = 1}^c \frac{1}{(\lambda + n_p)^2}\left(\sum_{i,j}k(\mathbf{x}_{(p),i}^{s}, \mathbf{x}_{(p),j}^{s})\right)  \\ 
      & + \sum_{p = 1}^c \frac{1}{(\lambda + n_p)^2}\left(\sum_{i,j}k(\mathbf{x}_{(p),i}^{t}, \mathbf{x}_{(p),j}^{t})\right)  \\ 
      & -2 \sum_{p = 1}^c \frac{1}{(\lambda + n_p)^2}\left(\sum_{i,j}k(\mathbf{x}_{(p),i}^{s}, \mathbf{x}_{(p),j}^{t})\right),   
    \end{aligned}
  \end{equation}
  where $n_i$ denotes the sample-size of $i$-th label such that $n = \sum_{i = 1}^{c}n_i$, $\mathbf{x}_{(p),i}$ denotes the $i$-th sample of $p$-th label.   
\end{proposition}
Eq.~\eqref{fuc:emp_CMMD} contains three terms. The third term characterizes the cross-domain intra-class similarity. When Eq.~\eqref{fuc:emp_CMMD} is minimized, the third term is maximized so that the cross-domain sample with the same label will be aligned, leading to conditional alignment as expected. However, the first two terms of the equation represent the intra-class similarity of both domains. Minimizing them will decrease the intra-class similarity, which means intra-class samples will be pushed away from each other and the discriminability of representations is degraded.
\begin{remark}
  Note the Kronechker Delta kernel is defined as 
  \begin{equation}
    \begin{aligned}
    k(\mathbf{y}_1, \mathbf{y}_2) = \begin{cases}1\text{ if }\mathbf{y}_1 = \mathbf{y}_2\\0\text{ otherwise}\end{cases}.
    \end{aligned} 
  \end{equation}
  When it comes to the Gaussian kernel for empirical modeling, the case of $\mathbf{y}_1 = \mathbf{y}_2$ is the same; for other cases, the kernel values can be arbitrarily close to zero, \ie, for any $\epsilon>0$, there always width $\sigma_0$, such that 
  \begin{equation}
      \begin{aligned}
        k(\mathbf{y}_1, \mathbf{y}_2) = \exp\left(-\|{\mathbf{y}_1}-{\mathbf{y}_2}\|_2^2/\sigma_0^2\right) < \epsilon.
      \end{aligned}
    \end{equation}
  Thus, it is generally reasonable to use Gaussian kernel, which admits a similar property to Kronechker Delta kernel, for empirical modeling.
\end{remark}

\begin{remark}
  Note that although we treat the number of distinct labels as finite numbers in mini-batch training, it does not mean class-wise matching can be conducted. In fact, even though the pseudo target label has finite range in mini-batch, it still admits the continuous property, which means we can not cluster samples as usually done in classification methods. When two samples have predicted labels with little numerical difference, they can be regarded as samples of the same class in classification settings. Nonetheless, this kind of clustering is infeasible in regression settings since the labels are continuous. Therefore, class-wise alignment is not practical even in mini-batch training.
\end{remark}

With these observation, it is reasonable to reformulate Eq.~\eqref{fuc:emp_CMMD} to promote the discriminability of representations: 
\begin{equation}
  \label{fuc:emp_CMMD_mod}
  \begin{aligned}
    \widehat{d}_\mathrm{CMMD_{mod}}  =  
    & - \mathrm{tr}(\mathbf{K}^{tt}_{Y}(\mathbf{K}^{tt}_{Y})^{-1}\mathbf{K}^{tt}_{X}(\mathbf{K}^{tt}_{Y})^{-1}) 
     - \mathrm{tr}(\mathbf{K}^{tt}_{Y}(\mathbf{K}^{tt}_{Y})^{-1}\mathbf{K}^{tt}_{X}(\mathbf{K}^{tt}_{Y})^{-1}) \\ 
    & -2\mathrm{tr}(\mathbf{K}_{Y^tY^s}(\mathbf{K}^{ss}_{Y})^{-1}\mathbf{K}_{X^sX^t}(\mathbf{K}^{tt}_{Y})^{-1}) 
  \end{aligned}
\end{equation}
Now minimizing Eq.~\eqref{fuc:emp_CMMD_mod} will maximize intra-class similarity regardless of domain, yielding conditional matching with great discriminability. Then the empirical COD is modified as  
\begin{equation}
  \label{fuc:emp_COD_mod}
  \begin{aligned}
    \widehat{d}_\mathrm{COD_{mod}} = 
    & - \mathrm{tr}(\mathbf{K}^{tt}_{Y}(\mathbf{K}^{tt}_{Y})^{-1}\mathbf{K}^{tt}_{X}(\mathbf{K}^{tt}_{Y})^{-1}) 
     - \mathrm{tr}(\mathbf{K}^{tt}_{Y}(\mathbf{K}^{tt}_{Y})^{-1}\mathbf{K}^{tt}_{X}(\mathbf{K}^{tt}_{Y})^{-1}) \\ 
    & -2\mathrm{tr}(\mathbf{K}_{Y^tY^s}(\mathbf{K}^{ss}_{Y})^{-1}\mathbf{K}_{X^sX^t}(\mathbf{K}^{tt}_{Y})^{-1}) \\ 
    & + \varepsilon\mathrm{tr}\left[{\mathbf{G}}_X^s\left(\varepsilon n{\mathbf{I}}_n+{\mathbf{G}}_Y^s\right)^{-1}\right]+\varepsilon\mathrm{tr}\left[\mathbf{G}_X^t\left(\varepsilon n{\mathbf{I}}_n+\mathbf{G}_Y^t\right)^{-1}\right] \\ 
    & - \frac2{n}\left\|\left(\mathbf{H}_n{\mathbf{A}}^t\right)^T{\mathbf{K}}^{ts}_{X}\left(\mathbf\mathbf{H}_n{\mathbf{A}}^s\right)\right\|_*
  \end{aligned}
\end{equation}

\subsection{COD-based DAR Modeling}
\label{sec:Overall}

According to Theorem~\ref{them:sufficient_con} , when the conditional distributions $P_{X^s\mid Y}$ and $P_{X^t\mid Y}$ are aligned, a well-trained source predictor can gain promising performance on target domain. However, when the shift is significant, it is difficult for predictor to produce high-quality pseudo labels at the beginning. For this reason, the model is likely to miss the optimal optimization direction in the early training stage and thus fall into sub-optimal solutions. Therefore, the marginal alignment can improve the reliability of initial pseudo labels, and further benefit the conditional alignment. Overall, the learning objective of COD can be formulated as 
\begin{align}
  \label{fuc:overall2}
  \min_{g,h}  ~ \mathcal{L}_\mathrm{src} + \lambda_1 \widehat{d}_\mathrm{COD_{mod}} + \lambda_2 \widehat{d}_\mathrm{KGW}, 
\end{align}
where $\lambda_1$, $\lambda_2$ are the trade-off parameters. For regression task, the source risk $\mathcal{L}_\mathrm{src}$ is commonly set as Mean Square Error (MSE) over continuous outputs:
\begin{align}
  \label{fuc:mse}
  \mathcal{L}_\mathrm{src} = \frac{1}{n^s}\sum_{i = 1}^{n^s} \left\lVert \hat{\mathbf{y}^s_i} - \mathbf{y}_i^s \right\rVert _2^2 
\end{align}
where $\hat{\mathbf{y}^s_i}$ is the predicted value of the $i$-th source sample. Note that the marginal alignment term can be replaced by any other discrepancy metric, while we adopt KGW here to preserve the consistency with OT framework.

\section{Experiments}
We evaluate our method with several state-of-the-art domain adaptation methods on three regression benchmark datasets, implementation details and additional illustrations are provided in supplementary material.

\textbf{dSprites}~\cite{higgins2016beta} is a standard 2D synthetic dataset for deep representation learning, which contains three domains each with 737,280 images: Color (\textbf{C}), Noisy (\textbf{N}) and Scream (\textbf{S}). Following~\cite{chen2021representation}, we consider three factors for regression tasks: scale and two plane coordinates, while the orientation factor is excluded for its complexity. We evaluate all methods on six transfer tasks: $\textbf{C} \to \textbf{N}$, $\textbf{C} \to \textbf{S}$, $\textbf{N} \to \textbf{C}$, $\textbf{N} \to \textbf{S}$, $\textbf{S} \to \textbf{C}$ and $\textbf{S} \to \textbf{N}$. Following previous works~\cite{chen2021representation,nejjar2023dare}, the sum of Mean Absolute Error (MAE) on all three sub-regression tasks is reported as the evaluation metric.

\textbf{Biwi kinect}~\cite{fanelli2013random} is a real-world dataset containing two domains according to gender: Female (\textbf{F}) with 5874 images and Male (\textbf{M}) with 9804 images. The three factors that can be employed for regression tasks are pitch, yaw and roll. We evaluate all methods on two transfer tasks: $\textbf{F} \to \textbf{M}$ and $\textbf{M} \to \textbf{F}$. The sum of MAE on three sub-regression tasks is reported.

\textbf{MPI3D}~\cite{gondal2019transfer} is a simulation-to-real dataset of 3D objects which consist of three different domain: Toy (\textbf{T}), RealistiC (\textbf{RC}) and ReaL (\textbf{RL}). Each domain contains 1,036,800 images. There are two factors in MPI3D that can be employed for regression tasks: a rotation about a vertical axis and a second rotation about a horizontal axis. We evaluate all methods on six transfer tasks: $\textbf{RL} \to \textbf{RC}$, $\textbf{RL} \to \textbf{T}$, $\textbf{RC} \to \textbf{T}$, $\textbf{RC} \to \textbf{RL}$, $\textbf{T} \to \textbf{RL}$ and $\textbf{T} \to \textbf{RC}$. The sum of MAE on two sub-regression tasks is reported.

\textbf{Comparison methods}. SOTA methods for UDA and DAR are selected: TCA~\cite{pan2010domain}, MCD~\cite{saito2018maximum}, DAN~\cite{long2015learning}, DANN~\cite{ganin2016domain}, JDOT~\cite{courty2017joint}, AFN~\cite{xu2019larger}, RSD~\cite{chen2021representation}, DARE-GRAM~\cite{nejjar2023dare}. Following previous works~\cite{chen2021representation,nejjar2023dare}, Mean Absolute Error (MAE) is used as our evaluation metric across all regression tasks. In comparison experiment, the proposed DAR model in Eq.~\eqref{fuc:overall2} is denoted as COD for convenience.

\subsection{Results and Analysis}
\textbf{Comparison on dSprites.} 
As shown in~\cref{tab:result_dSprites}, COD achieves the best performance among all competing methods in nearly all tasks of dSprites. In $\textbf{C} \to \textbf{N}$ and $\textbf{C} \to \textbf{S}$, in which the Resnet-18 and UDA methods exhibit terrible performance because of the significant domain gap. Although SOTA DAR methods have greatly boosted the performance compared with classification methods, COD still achieves a significant improvement, reducing the error to only $40\%$ of that produced by SOTA methods. These results demonstrate that the conditional invariant representations learned by COD indeed ensure better discriminability.

\noindent
\textbf{Comparison on Biwi Kinect.}
Since Biwi Kinect is the closest to real-world scenarios among the three benchmark datasets, the performance improvement of model is more challenging. In~\cref{{tab:result_BiwiKinect}}, the performance gap between UDA methods and DAR methods is relatively smaller, which implies the SOTA DAR methods with marginal alignment indeed show limitations in the difficult transfer tasks. In such a scenario, the sufficiency of condition alignment is directly validated, where the MAE of COD is significantly lower than other SOTA methods. Thus, for real-world regression applications with complex shift, the COD model indeed shows superiority in both theoretical and methodological aspects.

\noindent
\textbf{Comparison on MPI3D.}
In MPI3D, the resolution remains the same with dSprites, while the main challenge is the large sample size. In \cref{tab:result_MPI3D}, the MAE of COD is competitive with the SOTA method DARE-GRAM in the first four tasks, while higher in the last two tasks. However, note that COD still achieves the second-best result in averaged MAE, and obtains the lowest MAE in task RL$\to$T. In conclusion, the overall MAE of COD on three benchmarks is still lower, which demonstrates COD is generally superior to SOTA methods.

\begin{table}[t]
\renewcommand\tabcolsep{0.6pc}
\renewcommand{\arraystretch}{0.95}
  \caption{Domain adaptation regression results (MAE) on \textbf{dSprites} (ResNet-18).}
  \label{tab:result_dSprites}
  \centering
  \begin{tabular}{@{}lccccccc@{}}
    \toprule
     \textbf{Method} & C$\to$ N & C$\to$ S & N$\to$ C & N$\to$ S & S$\to$ C & S$\to$ N & \textbf{Avg.} \\
    \midrule
    Resnet-18 & 0.94 & 0.90 & 0.16 & 0.65 & 0.08 & 0.26 & 0.498 \\ 
    TCA & 0.94 & 0.87 & 0.19 & 0.66 & 0.10 & 0.23 & 0.498 \\ 
    MCD & 0.81 & 0.81 & 0.17 & 0.65 & 0.07 & 0.19 & 0.450 \\ 
    JDOT& 0.86 & 0.79 & 0.19 & 0.64 & 0.10 & 0.23 & 0.468 \\ 
    AFN & 1.00 & 0.96 & 0.16 & 0.62 & 0.08 & 0.32 & 0.523 \\ 
    DAN & 0.70 & 0.77 & 0.12 & 0.50 & 0.06 & 0.11 & 0.377 \\ 
    DANN& 0.47 & 0.46 & 0.16 & 0.65 & \textbf{0.05} & 0.10 & 0.315 \\ 
    RSD & 0.31 & 0.31 & 0.12 & 0.53 & 0.07 & 0.08 & 0.237 \\   
    DARE-GRAM & 0.30 & 0.20 & 0.11 & 0.25 & \textbf{0.05} & \textbf{0.07} & 0.164 \\ 
    \hline
    \rowcolor[gray]{.9}
    \textbf{COD} & \textbf{0.12} & \textbf{0.16} & \textbf{0.10} & \textbf{0.23} & 0.08 & 0.12 & \textbf{0.134} \\ 
    \bottomrule
  \end{tabular}
\end{table}

\begin{table}[t]
  \renewcommand\tabcolsep{0.6pc}
  \renewcommand{\arraystretch}{0.95}
  \caption{Domain adaptation regression results (MAE) on \textbf{Biwi Kinect} (ResNet-18).}
  \label{tab:result_BiwiKinect}
  \centering
  \begin{tabular}{@{}lccc@{}}
      \toprule
      \textbf{Method} & M$\to$ F & F$\to$ M & \textbf{Avg.} \\ 
      \midrule 
      Resnet-18 & 0.29 & 0.38 & 0.335 \\ 
      TCA  & 0.31 & 0.39 & 0.350 \\ 
      MCD  & 0.31 & 0.37 & 0.340 \\ 
      JDOT & 0.29 & 0.39 & 0.340 \\ 
      AFN  & 0.32 & 0.41 & 0.365 \\ 
      DAN  & 0.28 & 0.37 & 0.325 \\ 
      DANN & 0.30 & 0.37 & 0.335 \\ 
      RSD  & 0.26 & 0.30 & 0.280 \\ 
      DARE-GRAM & 0.23 & 0.29 & 0.260 \\ 
      \hline
      \rowcolor[gray]{.9}
      \textbf{COD} & \textbf{0.20} & \textbf{0.21} & \textbf{0.205} \\ 
      \bottomrule
  \end{tabular}
\end{table}

\begin{table}[tb]
\renewcommand\tabcolsep{0.36pc}
\renewcommand{\arraystretch}{0.95}
  \caption{Domain adaptation regression results (MAE) on \textbf{MPI3D} (ResNet-18).}
  \label{tab:result_MPI3D}
  \centering
  \begin{tabular}{@{}lccccccc@{}}
    \toprule
     \textbf{Method} & RL$\to$ RC & RL$\to$ T & RC$\to$ RL & RC$\to$ T & T$\to$ RL & T$\to$ RC & \textbf{Avg.} \\ 
    \midrule
    Resnet-18 & 0.17 & 0.44 & 0.19 & 0.45 & 0.51 & 0.50 & 0.377 \\ 
    TCA  & 0.17 & 0.42 & 0.19 & 0.42 & 0.50 & 0.50 & 0.373 \\ 
    MCD  & 0.13 & 0.40 & 0.15 & 0.45 & 0.52 & 0.50 & 0.358 \\ 
    JDOT & 0.16 & 0.41 & 0.16 & 0.41 & 0.47 & 0.47 & 0.353 \\ 
    AFN  & 0.18 & 0.45 & 0.20 & 0.46 & 0.53 & 0.53 & 0.390 \\ 
    DAN  & 0.12 & 0.35 & 0.12 & 0.27 & 0.40 & 0.41 & 0.278 \\ 
    DANN & \textbf{0.09} & 0.24 & 0.11 & 0.41 & 0.48 & 0.37 & 0.283 \\ 
    RSD  & \textbf{0.09} & 0.19 & \textbf{0.08} & 0.15 & 0.36 & 0.36 & 0.205 \\ 
    DARE-GRAM & \textbf{0.09} & 0.15 & 0.10 & \textbf{0.14} & \textbf{0.24} & \textbf{0.24} & \textbf{0.160} \\    
    \hline    
    \rowcolor[gray]{.9}
    \textbf{COD} & 0.10 & \textbf{0.11} & 0.13 & 0.15 & 0.37 & 0.29 & 0.192 \\ 
    \bottomrule
  \end{tabular}
\end{table}

\begin{table}[t]
  \renewcommand\tabcolsep{0.35pc}
  \renewcommand{\arraystretch}{0.95}
  \caption{Ablation study on different metrics for DAR.}
  \label{tab:result_Alba}
  \centering
  \begin{tabular}{@{}cccc|cccccc@{}}
    \toprule
     \multicolumn{4}{c|}{\textbf{Objectives}}& \multicolumn{2}{c}{\textbf{dSprites}} &  \multicolumn{2}{c}{\textbf{Biwi Kinect}} & \multicolumn{2}{c}{\textbf{MPI3D}}  \\
     MSE & $d_\mathrm{KGW}$ & $d_\mathrm{COD}$ & ${d}_{\mathrm{COD}_{mod}}$ & C$\to$ S & N$\to$ S  & T$\to$ RL & T$\to$ RC & M$\to$ F & F$\to$ M \\ 
    \midrule
    $\checkmark$ & ~ & ~ & ~ & 0.90 & 0.65 & 0.29 & 0.38 & 0.51 & 0.50  \\ 
    $\checkmark$ & $\checkmark$ & ~ & ~ & 0.18 & 0.26 & 0.21 & 0.25 & 0.41 & 0.34  \\ 
    $\checkmark$ & ~ & $\checkmark$ & ~ & 0.26 & 0.45 & 0.21 & 0.22 & 0.40 & 0.31  \\ 
    $\checkmark$ & $\checkmark$ & $\checkmark$ & ~ & 0.17 & 0.25 & 0.21 & 0.22 & 0.41 & \textbf{0.29}   \\ 
    $\checkmark$ & $\checkmark$ & ~ & $\checkmark$ & \textbf{0.16} & \textbf{0.23} & \textbf{0.20} & \textbf{0.21} & \textbf{0.37} & \textbf{0.29}  \\ 
    \bottomrule
  \end{tabular}
\end{table}

\noindent
\textbf{Ablation study.}
The ablation experiment was conducted on the most challenging tasks from all three datasets:
$\textbf{C} \to \textbf{S}$ and $\textbf{N} \to \textbf{S}$ from dSprites, $\textbf{T} \to \textbf{RL}$ and $\textbf{T} \to \textbf{RC}$ from MPI3D, and both $\textbf{M} \to \textbf{F}$ and $\textbf{F} \to \textbf{M}$ from Biwi Kinect. The major components of proposed method, \ie, MSE objective $\mathcal{L}_{src}$, KGW distance $d_\mathrm{KGW}$, COD distance $d_\mathrm{COD}$ and modified COD ${d}_{\mathrm{COD}_{mod}}$, are evaluated, where the results are shown in~\cref{tab:result_Alba}.

From the 2\textsuperscript{nd} and 3\textsuperscript{rd} rows, it can be observed that both OT-based metrics show promising results on DAR datasets, where the conditional alignment with COD shows better performance than marginal alignment with KGW on MPI3D and Biwi Kinect datasets. On the other hand, since dSprites induce more difficult transfer problems, the baseline performance is generally lower, and the pseudo labels for conditional alignment are less reliable. Therefore, the combination of KGW and COD is necessary to ensure higher performance on challenging tasks, where the results in the 4\textsuperscript{th} row validate the mutually beneficial relation between these two objectives. Finally, the 5\textsuperscript{th} row shows the effectiveness of rethinking the discriminability of COD, which implies the modifications on the moment statistics indeed ensure lower MAE values on all datasets. Overall, these results demonstrate that conditional alignment is indeed crucial for successful DAR, and the theoretical results and proposed method are generally valid.

\begin{figure}[t]
  \centering
  \includegraphics[width=1.0\linewidth]{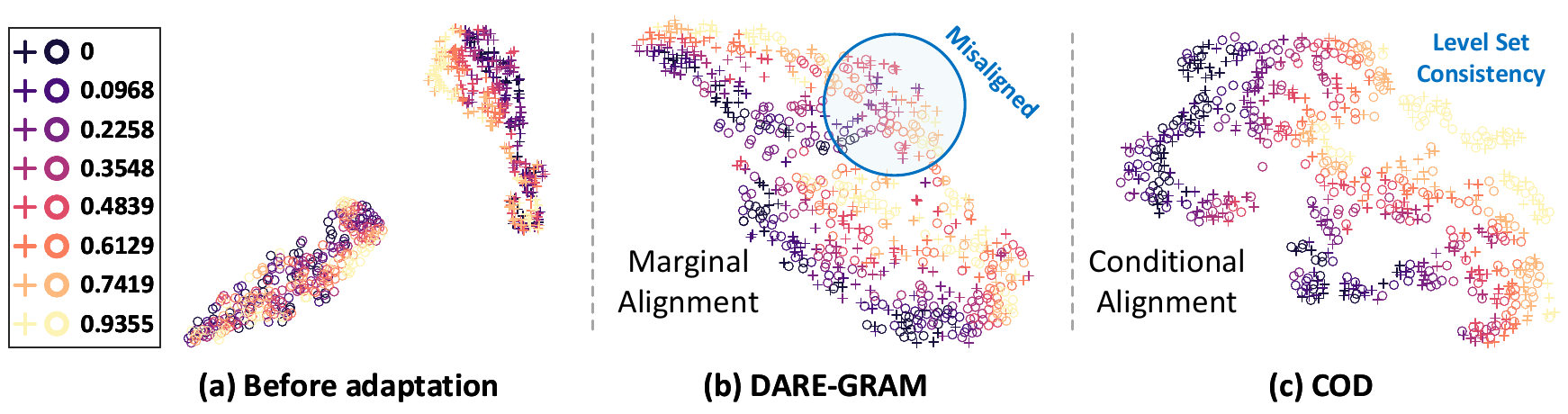}
  \caption{t-SNE visualization of learned representations. Label values are denoted by color gradients. Eight label values are selected from the range of the variable for visualization. '$+$': source samples; '$\circ$': target samples.}
  \label{fig:tsne}
\end{figure}

\noindent
\textbf{Feature Visualization.}
t-SNE~\cite{van2008visualizing} is employed to visualize the 2-D features of different alignments on dSprites C$\to$N task. \textbf{(a)} Before adaptation, distributions of source and target domain differ severely, as shown in~\cref{fig:tsne}(a), which implies predictor is infeasible for the target samples with a significant domain gap. \textbf{(b)} After marginal distribution alignment via the SOTA DARE-GRAM method, source and target representations are aligned globally, and the cross-domain distributions are overlapped on the whole. However, there are still misaligned local patterns where source and target samples of different colors are matched. Therefore, it is hard to fit clear level sets in~\cref{fig:tsne} (b). \textbf{(c)} After conditional distribution alignment via COD, source and target distributions are aligned not only globally but also locally. As shown in~\cref{fig:tsne} (c), representations are matched according to label values so the change of color shows a clear gradient and level sets suitable for both domains can be fitted. 

\begin{wrapfigure}{r}{0.4\linewidth}
\vskip -0.4in
\includegraphics[width=1\linewidth]{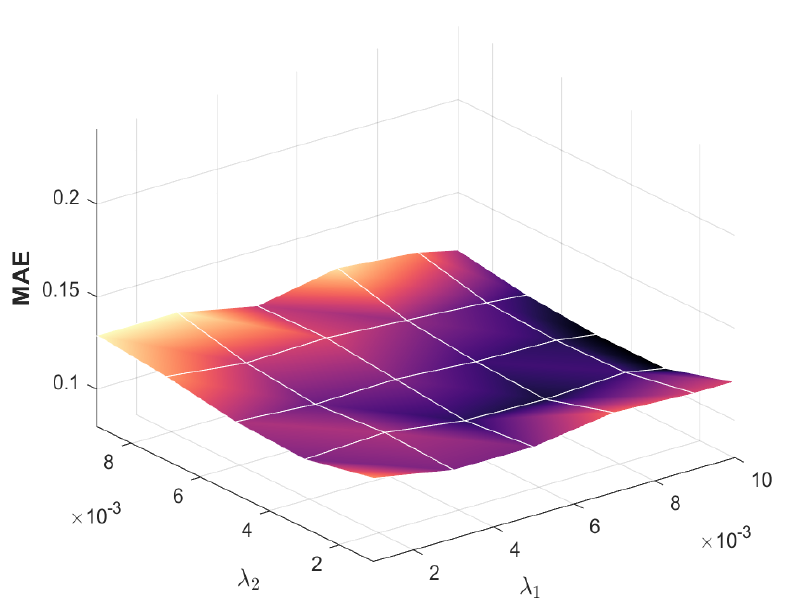}
\caption{MAE values under different settings of hyper-parameters.}
\label{fig:para}
\vskip -0.25in
\end{wrapfigure}

\noindent
\textbf{Hyper-parameter.}
We investigate the sensitivity of hyper-parameters on dSprites C$\to$N. MAE values by varying $\lambda_1$ and $\lambda_2$, \ie, parameters of modified COD and KGW, are shown in~\cref{fig:para}. The results show that the performance of the COD-based DAR model is generally robust to the different choice of parameters. In fact, the MAE values of the COD-based model are stable in random experiments, which implies the proposed method is both robust from the view of randomness and parameter setting. In conclusion, the stable performance validates that the proposed model is also empirically reliable.

\section{Conclusion}
\label{sec:conclusion}
In this work, we studied the limitations of existing DAR research, \ie, the misaligned intrinsic discriminant structure and negative transfer of marginal distribution matching framework. To deal with the challenges of conditional shift with continuous output space, we provide feasible solutions from both theoretical and methodological views, where the main results show that: 1) the cross-domain generalization regression error can be sufficiently bounded by the conditional discrepancy on continuous labels; 2) the proposed COD serves as a well-defined metric for conditional discrepancy measures and empirical modeling. With these guarantees, a COD-based DAR method is proposed with a discriminability enhancement mechanism. Numerical validations on standard DAR datasets validate the superiority of COD. 


\section*{Acknowledgements}
This work is supported in part by National Natural Science Foundation of China (Grant No. 62376291), in part by Guangdong Basic and Applied Basic Research Foundation (2023B1515020004), in part by Science and Technology Program of Guangzhou (2024A04J6413), in part by the Fundamental Research Funds for the Central Universities, Sun Yat-sen University (24xkjc013), in part by Guangdong Province Key Laboratory of Computational Science at Sun Yat-sen University (2020B1212060032), and in part by Key Laboratory of Machine Intelligence and Advanced Computing, Ministry of Education.

%
%
\bibliographystyle{splncs04}
\bibliography{main}
\end{document}